\pdfoutput=1

\documentclass[11pt]{article}
\usepackage[final]{coling}
\usepackage{array}
\usepackage{times}
\usepackage{latexsym}
\usepackage{longtable}
\usepackage{verbatim}
\usepackage{array}    
\usepackage{graphicx} 

\usepackage{multirow}
\usepackage[T1]{fontenc}
\usepackage{makecell}  
\usepackage{booktabs}  
\usepackage[utf8]{inputenc}

\usepackage{microtype}

\usepackage{inconsolata}

\usepackage{graphicx}

\title{SentiXRL: An advanced large language Model  Framework for Multilingual Fine-Grained Emotion Classification in Complex Text Environment}

\author{
  \textbf{Jie Wang\textsuperscript{1,*}},
  \textbf{Yichen Wang\textsuperscript{2,*}},
  \textbf{Zhilin Zhang\textsuperscript{3}},
  \textbf{Jianhao Zeng\textsuperscript{4}},
  \textbf{Kaidi Wang\textsuperscript{5}},
  \textbf{Zhiyang Chen\textsuperscript{6,**}}
  \\
  \textsuperscript{1}Tongji University,
  \textsuperscript{2}Tsinghua University,
  \textsuperscript{3}New York University,\\
  \textsuperscript{4}Tianjin University,
  \textsuperscript{5}Soochow University,
  \textsuperscript{6}Westlake University
  \\
  \small{
    \textbf{*} Co-first authors: Jie Wang (first author), Yichen Wang (co-first, second author).}
    }

\begin{document}
\maketitle

\begin{abstract}

 With strong expressive capabilities in Large Language Models(LLMs), generative models effectively capture sentiment structures and deep semantics, however, challenges remain in fine-grained sentiment classification across multi-lingual and complex contexts. To address this, we propose the Sentiment Cross-Lingual Recognition and Logic Framework (SentiXRL), which incorporates two modules,an emotion retrieval enhancement module to improve sentiment classification accuracy in complex contexts through historical dialogue and logical reasoning,and a self-circulating analysis negotiation mechanism (SANM)to facilitates autonomous decision-making within a single model for classification tasks.We have validated SentiXRL's superiority on multiple standard datasets, outperforming existing models on CPED and CH-SIMS,and achieving overall better performance on MELD,Emorynlp and IEMOCAP. Notably, we unified labels across several fine-grained sentiment annotation datasets and conducted category confusion experiments, revealing challenges and impacts of class imbalance in standard datasets. 

\end{abstract}

\section{Introduction}

Currently, utilizing large language models (LLMs) for text classification tasks is a prominent research focus\cite{brown2020language}. Specifically, text sentiment classification has garnered widespread attention due to its significance in understanding the nuances of human communication. Generative models, with their powerful expressive capabilities, can effectively capture the structure and deep semantics of emotional texts, thereby demonstrating outstanding performance in sentiment recognition and classification tasks,which is a solid foundation for other tasks such as roleplay, dialogue generation, and targeted content recommendations.Additionally, instruction fine-tuning of LLMs has proven their exceptional adaptability to various tasks\cite{ouyang2022training}. However, for more complex tasks such as fine-grained sentiment recognition, efficient processing frameworks are often required. In the context of multilingual communication and cultural differences, the complexity of multilingual understanding and response poses higher demands on the generalization capability of LLMs. The differing grammatical and syntactic features across languages, along with the limitations of traditional algorithms that focus on structured and short dialogue scenarios while overlooking more personalized user expressions, are among the many challenges that LLMs currently face.

Our goal is to design an efficient framework for fine-grained emotion classification tasks for LLMs in multilingual and complex text environment. To this end, we design the SentiXRL cross-lingual emotion recognition framework, which enables fine-grained emotion recognition in more complex textual environment across multiple languages. Our architecture primarily includes an efficient emotion retrieval enhancement module, which connects contextual information through historical dialogues and implicit inference while performing emotion reasoning. Additionally, we design a Self-Analytical Negotiation Mechanism (SANM) to help LLMs perform emotion verification and logical reasoning, thereby improving emotion classification capabilities in complex texts and contexts.

We validate our approach on several standard  benchmark datasets, surpassing existing SOTA on most benchmarks. Furthermore, we construct the largest fine-grained  sentiment annotation dataset to date and conduct category confusion experiments , verifying the impact of category imbalance  on LLMs.

Finally, our contributions are summarized as follows:

\begin{itemize}
\item We propose a novel framework specifically designed for the task of cross-lingual fine-grained emotion recognition in large language models.
\item A novel Self-Analytical Negotiation Mechanism (SANM) is introduced, enhancing emotion recognition accuracy in complex environments through logical reasoning and emotion verification.
\item SentiXRL outperforms most previous models on five standard Emotion Recognition in Conversations(ERC) benchmarks and achieves comprehensive single-modal state-of-the-art on two emotion analysis datasets.
\item To address category imbalance in mainstream datasets, we standardized label mapping across multiple fine-grained emotion annotation datasets and conducted category confusion experiments. Additionally, ablation studies on the ERC datasets highlight the advantages of the SANM module.
\end{itemize}

\section{Related Work}

Dialogue emotion recognition has evolved from traditional machine learning methods such as SVM, which focused primarily on general textual sentiment, to deep learning approaches. Notable among these are mainstream discriminative architectures like RNN, GNN, and LSTM \cite{2017Context}, which capture complex inter-sentence dependencies, or Transformers \cite{liu2023hierarchical} that effectively capture contextual information. These advancements have significantly enhanced the accuracy of analysis. The emergence of multimodal fusion (e.g., combining speech or facial emotions) has enabled these discriminative models to comprehensively understand and recognize emotional states in dialogues. However, the integration of more modalities introduces limitations in application scope and complexity in data collection. Consequently, some researchers have begun incorporating dialogue modeling and situational interactions \cite{lei2024instructerc} or attempting to infuse common-sense information into emotion recognition tasks \cite{yi2022contextual, Li2021PastPA}.

\begin{figure}[htbp]
  \centering
  \includegraphics[width=0.48\textwidth]{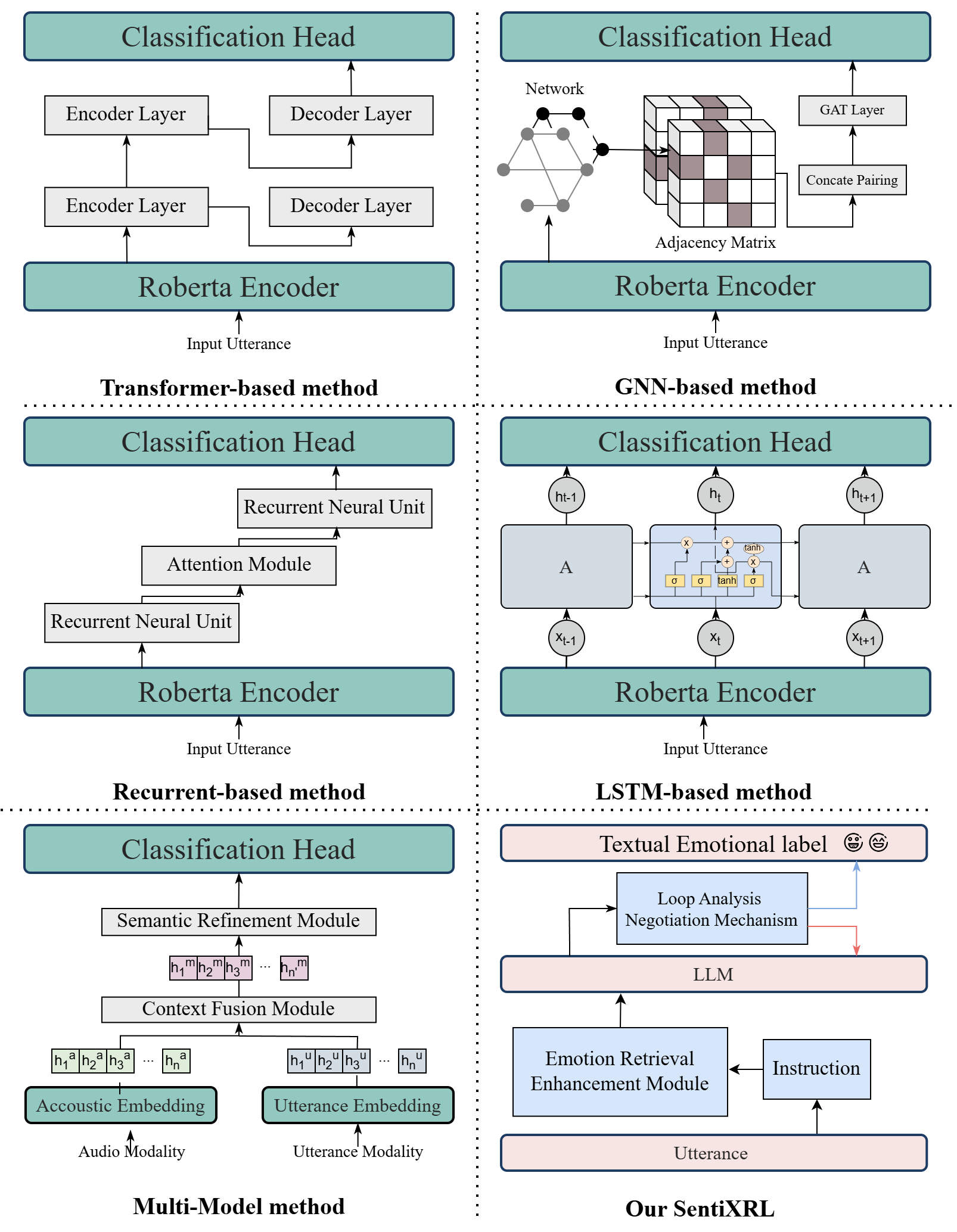}
  \caption{The illustration of different paradigms for ERC}
  \label{fig:example}
\end{figure}

\subsection{ Logical Reasoning in Text Emotion Recognition}
In our view, both common-sense information and other modalities serve as supplementary information external to the dialogue itself. These types of information cannot fully cover all dialogue scenarios or operate in constrained environment. Therefore, enabling models to reason and validate is the true solution for text emotion recognition tasks. Moreover, mainstream discriminative models suffer from complex system design and overfitting to specific datasets or dialogue patterns. Thus, generative architectures based on large language models (LLMs) have emerged as a novel approach to addressing these issues. The successful application and emergent capabilities of LLMs \cite{zhao2023survey} have demonstrated their excellent performance in natural language reasoning tasks. Research has shown that LLMs can follow contextual information \cite{NEURIPS2020_1457c0d6} and comprehend natural language instructions \cite{mishra-etal-2022-cross, chung2022scaling}. However, LLMs still underperform in reasoning tasks compared to smaller models (e.g., fine-tuned BERT) \cite{lee2023large}, presenting challenges for the application of LLM-based logical reasoning in text emotion recognition tasks.

\section{Methodology}

This chapter provides an in-depth overview of the novel SentiXRL architecture, detailing its emotion retrieval enhancement module, self-circulating analysis negotiation mechanism, and emotion analysis tasks. It also thoroughly explains the experimental training and inference processes.

\subsection{Emotion Retrieval Enhancement Module}
To better leverage the reasoning capabilities of large language models (LLMs), we restructure the ERC task into a sequence format, fine-tuning the LLM. To adapt the LLM to the specific emotion recognition task at hand, we design an efficient emotion retrieval module. As shown in Figure 2, this module consists of instructions, a history window, label statements, and emotional deduction.

\textbf{Instructions $I$}: Define the specific task content and standardized format.

\textbf{History Window $H$}: Represents the round of historical dialogue information used to connect the previous sequence of words, specifically in the form of:
\begin{equation}
H=[h_1,h_2,\dots,h_m]
\end{equation}

\textbf{Labels $L$}: Restrict the model's output range, allowing it to output label categories $ld\in D$ within the label domain $D$\footnote{In traditional sentiment analysis, sentiment primarily includes the categories positive, negative, and neutral, and many sentiment datasets follow this convention. In fine-grained datasets, emotion categories can be mapped to sentiment using emotion dictionaries. However, the surprise shown in Figure 2 needs to be categorized based on the text content. The emotion categories appearing in Figure 2 are those used in the datasets listed below.}.


\textbf{Emotional Deduction $E$}: Utilize the reasoning capabilities of the generative model to infer possible scenarios $S$, characters $P$, and relationships $R$ based on historical dialogue and the current statement. Thus, $E=(S,P,R)$.

Therefore, the task of this module can simplify the processing of the input statement $u_i$ as follows:

\begin{equation}
T_i=[I_{u_i},H_{u_i},ld_{u_i},E_{u_i}]
\end{equation}

\begin{figure*}[htbp] 
\centering
\includegraphics[width=1\textwidth]{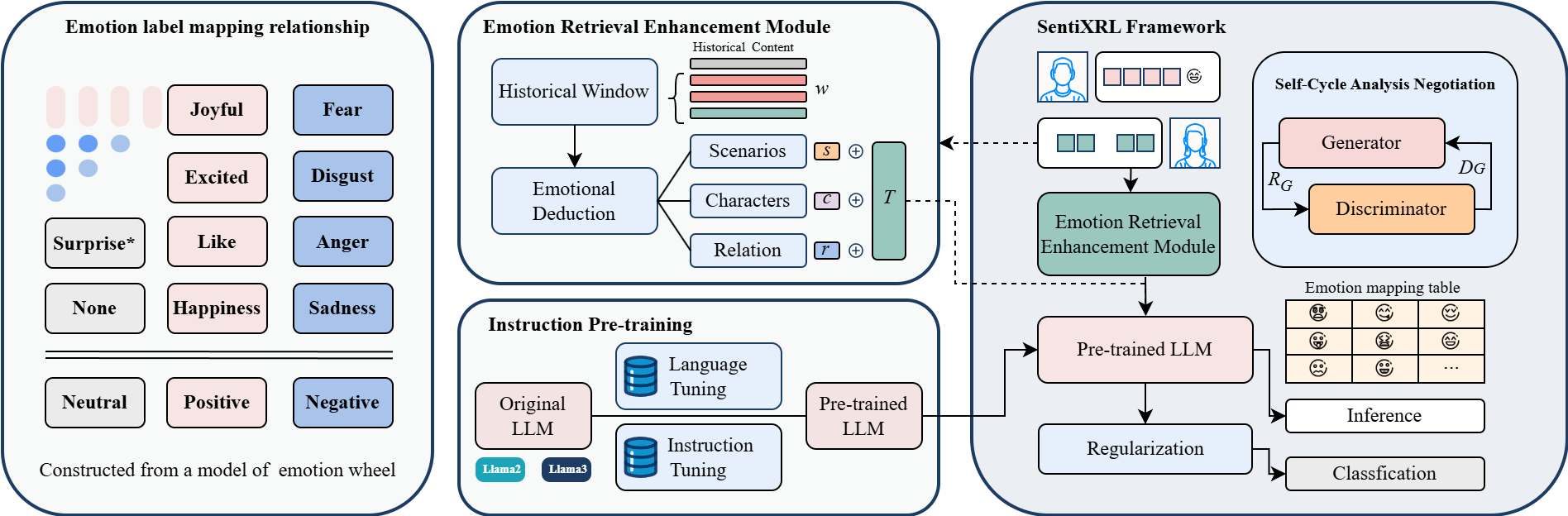}
\caption{Overview of SentiXRL Framework}
\label{fig:framework}
\end{figure*}

\subsection{Self-circular Analysis Negotiation Mechanism}

Due to the generative nature of LLM in sentiment analysis tasks, despite fine-tuning efforts and instruction-based constructions, there is still no guarantee that the output belongs to the specified sentiment category, especially for fine-grained sentiment classification tasks. To address the need for correction and supervision, and to mitigate the potential inaccuracies and lack of specificity in individual LLM outputs, WE propose a cyclic verification analysis negotiation mechanism to assist in completing sentiment analysis tasks. This mechanism differs from multi-LLM negotiation strategies and the supervised learning strategy using contextual learning (ICL) paradigms. 

The core of this strategy is a generative-discriminative architecture. Unlike conventional approaches requiring an additional discriminative model for supervision, the proposed approach leverages the original LLM model. Given the effectiveness of full parameter training of base models in fine-tuning tasks, there is no need for a separate discriminator. Instead, a new module framework is designed for the original model, integrating tasks of inference and output credibility. This allows for verification of classification objectives and explanatory derivation following structured reasoning chains after preliminary classification tasks.
\subsubsection{Inference Generator and Deduction Discriminator}
Compared to introducing a deduction discriminator, utilizing the distributional inference of two large language models makes the single LLM analysis negotiation framework more convenient and efficient in handling subtasks. The key lies in implementing an alternating role mechanism where a single LLM can function both as a generator and a discriminator. This paper defines distinct task templates for the generator and discriminator to ensure that the LLM comprehends its current role and task accurately. The generator judges the sentiment of text and generates a chain of distributional inferences and sentiment decisions, while the discriminator evaluates the generator's output and provides explanations.

\subsubsection{Role Alternation and Consensus Mechanism}
In each interaction round, the role of the LLM is clearly defined, and operations are performed according to different task templates through multi-round interaction processes. If the LLM reaches the same sentiment decision in two consecutive rounds, consensus is achieved. In case of disagreement, multiple attempts are made within the maximum round limit. If consensus is achieved, the process concludes; otherwise, an outlier is directly outputted if no valid sentiment decision is reached within the maximum round limit.

\subsubsection{Mathematical Modeling}
The self-circular analysis negotiation mechanism can also be represented mathematically by modeling the interaction between the generator and discriminator as an iterative process, defined as follows:

\begin{itemize}
  \item $G$: Generator
  \item $D$: Discriminator
  \item $T$: Sentiment analysis text
  \item $R_G$: Generator response (including sentiment analysis)
  \item $R_D$: Discriminator response (evaluating generator response)
  \item $E$: Final sentiment analysis
  \item $N$: Maximum number of cycles
\end{itemize}

The generator $G$ generates a response $R_G$ based on the sentiment analysis text $T$:
\begin{equation}
  R_G^{(n)} = G(T)
\end{equation}

The discriminator $D$ generates a response $R_D$ based on $R_G^{(n)}$:
\begin{equation}
  R_D^{(n)} = D(R_G^{(n)})
\end{equation}

If $R_D^{(n)}$ correctly evaluates the inference, $E = R_G^{(n)}$; otherwise, $n = n + 1$, repeating the above steps until $n = N$ or $R_D^{(n)}$ correctly evaluates the inference.

The self-circular analysis negotiation mechanism harnesses the capabilities of a single LLM across multiple attempts to achieve inference and decision-making. This method, compared to integrating a new inference generator utilizing the collective capabilities of two LLMs for the entire decision-making process, only requires a decision-making framework, enabling multi-LLM functionality. Moreover, this approach is applicable to other subtasks or LLM-based tasks, requiring only the redesign of the inference framework without the need for separate training of new discriminators.

Due to the imbalanced class distribution in the standard baseline dataset used, we adopt Focal Loss as the loss function for subsequent main tasks. This approach increases the weight of hard-to-classify samples, thereby mitigating the impact of class imbalance on the results:

\begin{equation}
L = -\alpha_t (1 - p_t)^\gamma \log(p_t)
\end{equation}

\begin{itemize}
\item $p_t$: The model's predicted probability for the actual class
\item $\alpha_t$: The weighting factor to balance positive and negative samples. For a sample of class $t$, if $y=1$, then $\alpha > 0$
\item $\gamma$: A modulating factor to reduce the weight of easy-to-classify samples
\end{itemize}

The core idea of Focal Loss is to introduce the modulating factor $(1 - p_t)^\gamma$, which reduces the loss contribution from samples that the model already predicts with high confidence, while increasing the loss contribution from samples that the model predicts with less confidence. This adjustment encourages the model to focus more on the hard-to-classify samples. In our experiments, the default parameters are set as: $\alpha = 0.25, \gamma = 2.0$.

\section{Experiments}

Initially, we considered using cross-lingual sentiment analysis datasets, such as XED\cite{ohman2020xed} or NaijaSenti\cite{muhammad2022naijasenti}, for our experiments. These datasets include texts from various language categories and offer fine-grained sentiment annotations. However, we ultimately decided against using them for the following reasons:

1)For the current task of dialog sentiment analysis, the most widely used sentiment datasets are primarily in English and Chinese. Many state-of-the-art methods and models have been validated on these datasets, making them more representative in testing. In contrast, multilingual datasets are still rarely used in this research area.

2)Taking XED as an example, most cross-lingual sentiment datasets are originally annotated in English, with annotations for other languages created through projection or translation. While this ensures linguistic diversity, the cultural and linguistic specificity of the original language limits the applicability of these annotations to other languages. This misalignment means that the dataset may not align with the usage habits of the target language communities, making it less representative for languages other than the one in which it was originally annotated.

Thus, we opted to focus on English and Chinese for the cross-lingual sentiment analysis task, using data collected from movies or personal daily content for training and testing, allowing the model to better adapt to real-world scenarios.


Firstly, for English emotion classification tasks, we selected several challenging datasets: MELD \cite{poria2018meld}, EmoryNLP \cite{zahiri2018emotion}, and IEMOCAP \cite{busso2008iemocap}. IEMOCAP is an interactive emotional dyadic motion capture database that covers emotional exchanges in daily life. The MELD and EmoryNLP datasets are derived from dialogues in the TV show \textbf{Friends}, providing contextual information and fine-grained emotion labels. Particularly, EmoryNLP also includes emotion annotations for long dialogue sequences. However, it is worth noting that all datasets have imbalanced emotion distributions.

\begin{figure}[htbp]
  \centering
  \includegraphics[width=0.48\textwidth]{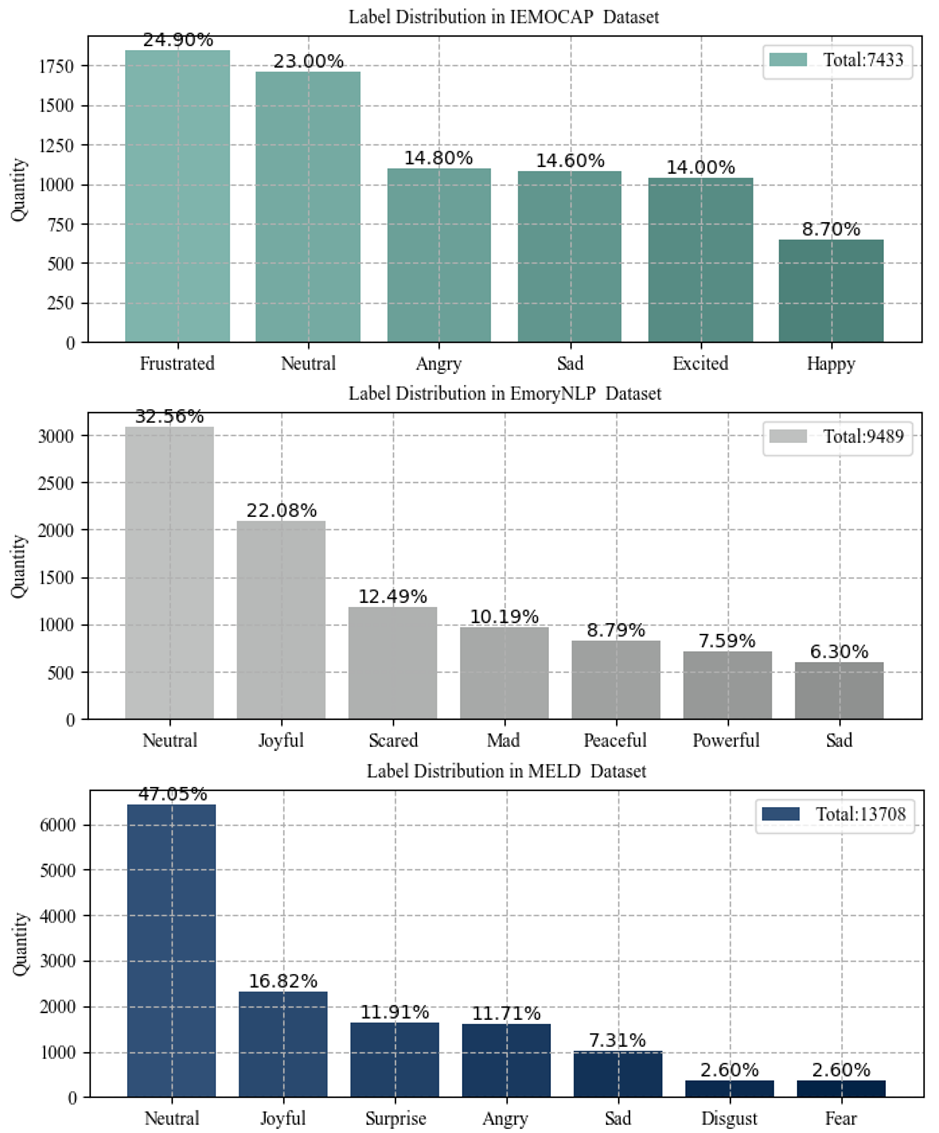}
  \caption{Label Distribution in MELD,EmoryNLP and IEMOCAP Dataset}
  \label{fig:example}
\end{figure}

For Chinese emotion recognition tasks, we chose CPED \cite{chen2022cped} and CH-SIMS \cite{liu2022make} as standard baseline datasets to evaluate the effectiveness of SentiXRL. The CPED dataset includes personal characteristics, various dialogue behaviors, and scenarios, while CH-SIMS offers richer character backgrounds and spans across different ages. Therefore, these datasets pose greater challenges. Although some of these datasets are multimodal, our study currently focuses solely on the emotion categories and textual modality of the data.

\begin{figure}[htbp]
  \centering
  \includegraphics[width=0.48\textwidth]{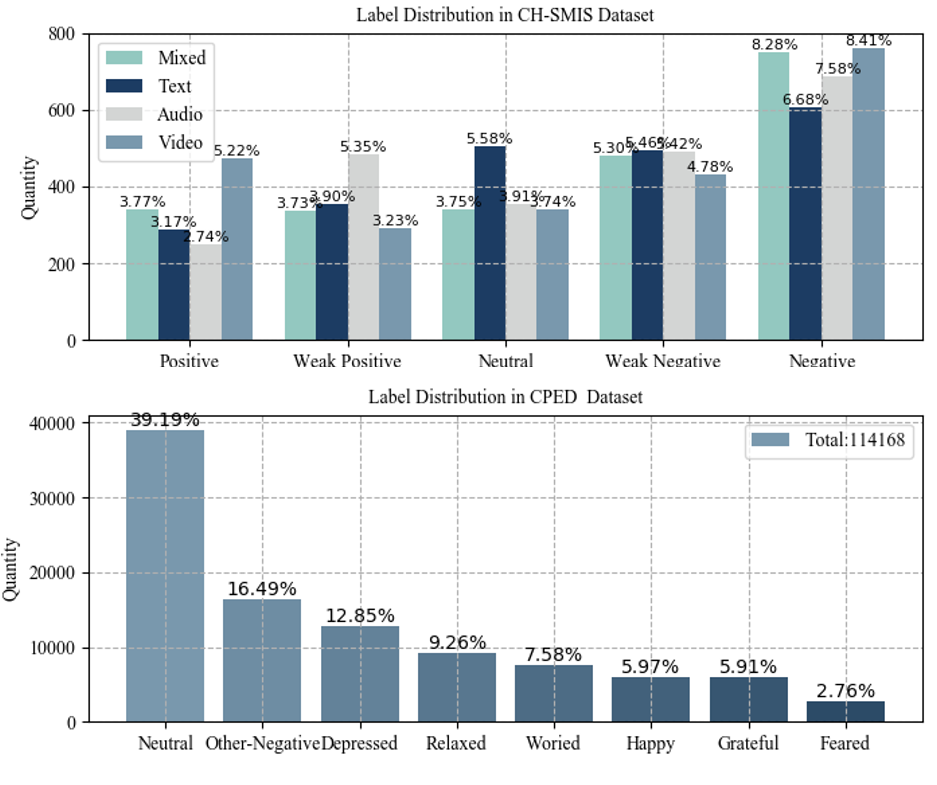}
  \caption{Label Distribution in CH-SIMS and CPED}
  \label{fig:example}
\end{figure}

We compared several single-text modality baselines with SentiXRL, conducting experiments on the Llama2-7B (L2) and Llama3-8B (L3) models. For Chinese emotion recognition tasks, the baseline models included MMML \cite{wumultimodal}, ALMT \cite{zhang2023learning}, bcLSTM \cite{poria2017context}, DialogXL \cite{shen2021dialogxl}, and BERT-AVG-MLP \cite{chen2022cped}. For English emotion recognition tasks, the baseline models consisted of SPCL+CL \cite{song2022supervised}, SACL \cite{hu2023supervised}, EmotionIC \cite{yingjian2023emotionic}, DualGATs \cite{zhang-etal-2023-pay}, and InstructERC \cite{lei2024instructerc}. For more detailed information on the baseline models and their implementations, please refer to the appendix.

\begin{table}[ht]
    \centering
    \caption{Results on two Chinese Benchmarks}
    \setlength{\tabcolsep}{3pt} 
    \renewcommand{\arraystretch}{1.2} 
    \small 
    \begin{tabular}{lcccc}
        \toprule
        \textbf{Dataset} & \multicolumn{2}{c}{\textbf{CH-SIMS}} & \multicolumn{2}{c}{\textbf{CPED}} \\
        \cmidrule(lr){2-3} \cmidrule(lr){4-5}
        \textbf{Models} & \textbf{F1} & \textbf{Accuracy} & \textbf{Macro-F1} & \textbf{Accuracy} \\
        \midrule
        \multicolumn{5}{c}{\textbf{Discriminant Models}} \\
        MMML & 82.9 & - & - & - \\
        ALMT & 81.57 & - & - & - \\
        bcLSTM & - & - & 49.65 & 45.40 \\
        DialogXL & - & - & 51.24 & 46.96 \\
        BERT-AVG-MLP & - & - & 51.50 & 48.02 \\
        \midrule 
        \multicolumn{5}{c}{\textbf{Generative Models}} \\
        SentiXRL($L_2$) & 76.51 & 83.15 & 32.96 & 47.00 \\
        SentiXRL($L_3$) & \textbf{82.83} & \textbf{84.20} & 45.31 & \textbf{50.70} \\
        \bottomrule
    \end{tabular}
\end{table}

\begin{table}[ht]
    \centering
    \caption{Results on three English Benchmarks}
    \setlength{\tabcolsep}{2.5pt} 
    \renewcommand{\arraystretch}{1.2} 
    \small 
    \begin{tabular}{lcccc}
        \toprule
        \textbf{Dataset} & \textbf{IEMOCAP} & \textbf{MELD} & \textbf{Emorynlp} & \textbf{Average} \\
        \textbf{Models} & \textbf{W-F1} & \textbf{W-F1} & \textbf{W-F1} & \textbf{W-F1} \\
        \midrule
        \multicolumn{5}{c}{\textbf{Discriminant Models}} \\
        \midrule
        SPCL+CL & \textbf{69.74} & 66.35 & 40.25 & 58.78 \\
        SACL & 69.22 & 66.45 & 39.65 & 58.44 \\
        EmotionIC & 69.61 & 66.40 & 40.01 & 58.67 \\
        DualGATs & 67.68 & \textbf{66.90} & \textbf{40.29} & 58.29 \\
        \midrule
        \multicolumn{5}{c}{\textbf{Generative Models}} \\
        \midrule
        InstructERC & \textbf{71.39} & \textbf{69.15} & 41.37 & 60.64 \\
        SentiXRL($L_2$) & 70.52 & 67.33 & 40.37 & 59.41 \\
        SentiXRL($L_3$) & 71.11 & 68.72 & \textbf{42.51} & \textbf{60.78} \\
        \bottomrule
    \end{tabular}
\end{table}

\subsection{Main Results}

Table 1 and Table 2 respectively present the comparative results of the SentiXRL model against other models on Chinese and English benchmark datasets. The experimental results indicate that our method significantly outperforms existing discriminative models and surpasses the current SOTA models in most benchmarks. Specifically, in the Chinese sentiment classification task, our accuracy on the CPED dataset shows an improvement of 5.6\% over the existing SOTA, and the F1 score on the CH-SIMS dataset increases by 1.55\%. Similarly, in the English sentiment classification benchmarks, SentiXRL achieves the highest individual performance on the more challenging EmoryNLP dataset and surpasses the existing SOTA in the average Weighted-F1 score across three datasets. The experimental results demonstrate that SentiXRL excels in both Chinese and English linguistic environment, validating the model's compatibility and adaptability in multilingual contexts.

\subsection{Ablation Study}

The ablation study results indicate that removing any component leads to a decline in relevant metrics, demonstrating that each part of the SentiXRL model is essential. Notably, the performance significantly drops when the ANM cyclic negotiation mechanism is removed, further proving the importance of this module. Additionally, multiple experiments have shown that this mechanism effectively helps the LLM focus on the current task. Even without fine-tuning, the model's task execution capability is relatively well improved.

\begin{table*}[ht]
    \centering
    \caption{The ablation results of Llama2 and Llama3 on five benchmarks}
    \setlength{\tabcolsep}{2pt} 
    \renewcommand{\arraystretch}{1.2} 
    \footnotesize 
    \begin{tabular}{lcccccccc}
        \toprule
        \textbf{Dataset} & \multicolumn{4}{c}{\textbf{IEMOCAP, MELD, EmoryNLP, Average }} & \multicolumn{2}{c}{\textbf{CH-SIMS}} & \multicolumn{2}{c}{\textbf{CPED}} \\
        \cmidrule(lr){2-5} \cmidrule(lr){6-7} \cmidrule(lr){8-9}
        \textbf{Models} & \textbf{Weighted-F1} & \textbf{Weighted-F1} & \textbf{Weighted-F1} & \textbf{Weighted-F1} & \textbf{F1} & \textbf{Accuracy} & \textbf{Macro-F1} & \textbf{Accuracy} \\
        \midrule
        \multicolumn{9}{c}{\textbf{Zero-shot+SentiXRL}} \\
        \midrule
        w/o $_{SANM}+L_2$ & - & - & - & - & 37.6 & 26.3 & 11.2 & 10.5 \\
        $L_2$ & 41.72 & 32.14 & 27.13 & 33.66 & 52.5 & 49.9 & 21.7 & 27.8 \\
        w/o $_{SANM}+L_3$ & - & - & - & - & 30.8 & 29.5 & 13.0 & 28.3 \\
        $L_3$ & 44.85 & 31.67 & 27.96 & 34.83 & 58.7 & 56.0 & 36.8 & 36.4 \\
        \midrule
        \multicolumn{9}{c}{\textbf{LoRA+Backbone}} \\
        \midrule
        w/o $_{SANM}+L_2$ & - & - & - & - & 59.3 & 59.8 & 24.3 & 42.7 \\
        $L_2$ & 53.27 & 37.69 & 29.51 & 40.16 & 61.2 & 63.6 & 26.8 & 44.5 \\
        w/o $_{SANM}+L_3$ & - & - & - & - & 65.0 & 68.3 & 37.5 & 47.7 \\
        $L_3$ & 55.01 & 38.08 & 30.28 & 41.12 &67.5 & 71.0 & 39.8 & 49.2 \\
        \midrule
        \multicolumn{9}{c}{\textbf{LoRA+SentiXRL}} \\
        \midrule
        w/o $_{SANM}+L_2$ & 70.52 & 67.33 & 40.37 & 59.41 & 76.5 & 83.2 & 33.0 & 47.0 \\
        w/o $_{SANM}+L_3$ & \textbf{71.11} & \textbf{68.72} & \textbf{42.51} & \textbf{60.78} & \textbf{82.8} & \textbf{84.2 }& \textbf{45.3} & \textbf{50.7} \\
        \bottomrule
    \end{tabular}
\end{table*}

\subsection{Category Impact Verification}
To validate the impact of data categories on the results of SentiXRL, we conduct an experiment addressing the inconsistency of emotion labels in most textual sentiment datasets. In this experiment, we collect high-quality, fine-grained Chinese textual sentiment classification datasets that are currently open-source on the Chinese internet. We standardize the emotion labels across all datasets (the mapping rules are shown in Figure 2) and perform data processing and cleaning. Detailed information on data processing and dataset composition can be found in Appendix B.

\begin{figure*}[htbp] 
\centering
\includegraphics[width=0.95\textwidth]{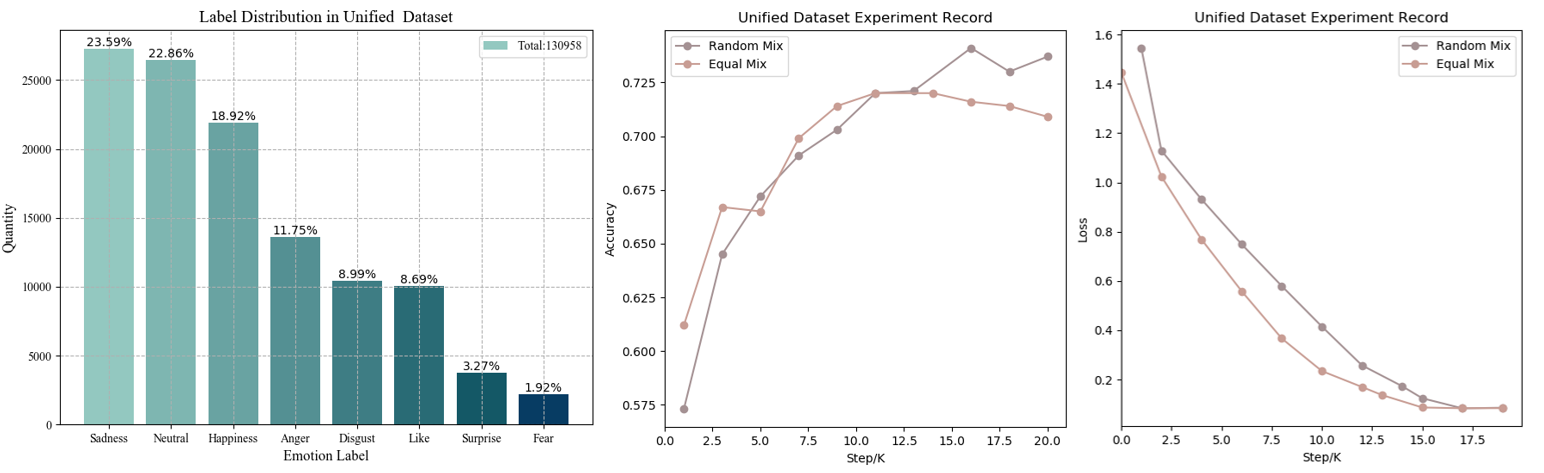}
\caption{Experimental results of SentiXRL($L_3$) on  Unified Dataset}
\label{fig:framework}
\end{figure*}

In the experiment, we design two classification methods: random data mixing and equal category mixing. These methods are intended to explore the impact of different data mixing strategies on the model. We hypothesize that due to the imbalance of categories in various datasets, the independent features of smaller datasets or those with fewer categories may be overshadowed by larger datasets. Therefore, equal category sampling can better highlight the characteristics and impacts of datasets with fewer categories. Both methods used the same data scale and hyperparameter settings.

Figure 5 presents the results of category validation. The experimental results indicate that, after 12K steps, the random mixing method surpasses the equal mixing method in  accuracy. This demonstrates SentiXRL's advantage in recognizing certain emotional categories. For more complex emotional categories, such as surprise, the limited textual content makes recognition relatively more challenging, but this result aligns with expectations. Additionally, the loss graph shows that the dataset with equal mixing converges more rapidly. Due to the balanced distribution of categories, the model can learn the feature mapping relationships for all categories more quickly. Overall, data category impact on SentiXRL's accuracy is minimal.

\begin{table*}[htbp] 
    \centering 
    \caption{The Main results on Twitter Benchmark}
    \label{my-label} 
    \begin{tabular}{lcccccc} 
        \toprule
        \textbf{Models} & \textbf{Venue} & \textbf{Acc(15)} & \textbf{Mac-F1} & \textbf{Acc(17)} & \textbf{Mac-F1} \\
        \midrule
        \multicolumn{6}{c}{\textbf{Text Only}} \\
        \cmidrule(lr){1-6}
        AE-LSTM\cite{wang2016attention} & EMNLP 2016 & 70.30 & 63.43 & 61.67 & 57.97 \\
        MemNet\cite{tai2017memnet} & EMNLP 2016 & 70.11 & 61.76 & 64.18 & 60.90  \\
        RAM\cite{zhang2023recognize} & EMNLP 2017 & 70.68 & 63.05 & 64.42 & 61.01  \\
        MGAN\cite{2018MGAN} & EMNLP 2018 & 71.17 & 64.21 & 64.75 & 61.46  \\
        BERT\cite{devlin2019bert} & NAACL 2019 & 74.15 & 68.86 & 68.15 & 65.23  \\
        \midrule
        
        \multicolumn{6}{c}{\textbf{Image and Text}} \\
        \cmidrule(lr){1-6}
        MIMN\cite{xu2019multi} & AAAI 2019 & 71.84 & 65.59 & 65.88 & 62.99 \\
        ESAFN\cite{2020Entity} & TASLP 2019 & 73.38 & 67.37 & 67.83 & 64.22 \\
        VilBERT\cite{lu2019vilbert} & NeurIPS 2019 & 73.69 & 69.53 & 67.86 & 64.93  \\
        TomRoBERTa\cite{2019Adapting} & IJCAI & 77.46 & 72.95 & 71.12 & 69.49  \\
        ModalNet-Bert\cite{2021ModalNet} & WWW 2021 & 76.71 & 70.93 & 69.55 & 67.28  \\
        EF-CapRoBERTa\cite{2021Exploiting} &  ACM  & 78.19 & 73.51 & 71.14 & 68.74  \\
        FITE\cite{yang-etal-2022-face} & EMNLP 2022 & 78.49 & 73.90 & 70.90 & 68.70  \\
        HIMT\cite{9765342} & TAC 2022 & 78.14 & 73.68 & 71.14 & 69.16  \\
        ITM\cite{ijcai2022p622} & IJCAI 2022 & 78.27 & 74.19 & 72.61 & 71.97  \\
        \midrule
        
        \multicolumn{6}{c}{\textbf{Text Only}} \\
        \cmidrule(lr){1-6}
        Ours & - & 77.28 & 70.93 & 70.84 & 69.12 \\
        \bottomrule
    \end{tabular}
    
\end{table*}

\subsection{Model performance validation in complex text environments}

To further validate the generalization capability of the SentiXRL model, this study selected two additional datasets, Twitter2015\cite{liu2015real} and Twitter2017\cite{rosenthal-etal-2017-semeval}, used for sentiment analysis and assessing information veracity. In contrast to the datasets used in the main experimental section, Twitter2015 and Twitter2017 feature longer text lengths. Unlike CPED, MELD, and EmoryNLP datasets sourced from scripted and emotional short texts from various film and TV show dialogues, the Twitter datasets originate from social media, where data is more heterogeneous and topics are less defined, thereby posing a greater challenge to the model's recognition abilities and subjecting its performance to more rigorous testing in complex textual and linguistic environments. We conducted comparative experiments and evaluated SentiXRL's performance on these datasets using Macro-F1 and Accuracy metrics.

The experimental results indicate that SentiXRL outperforms existing models in single-modal text analysis, particularly on the Twitter2015 and Twitter2017 datasets\footnote{The abbreviations `15` and `17` in the `Acc` and `Mac-F1` columns refer to Twitter 2015 and Twitter 2017, respectively.}. It achieved improvements of 2.6\% and 2.87\% in Accuracy, and 3.21\% and 4.63\% in Macro-F1 scores, respectively. Additionally, SentiXRL's performance in multimodal models also ranks among the top. Therefore, these results confirm that SentiXRL performs exceptionally well across both standard datasets and datasets characterized by significant information noise and unclear themes.

\section{Conclusion}
We introduce the SentiXRL model — an advanced large language model framework for multilingual fine-grained emotion classification in complex text environment,and discuss the Emotion Retrieval Enhancement Module and Self-circular Analysis Negotiation Mechanism within this architecture. This study utilized bilingual Chinese-English datasets and validated the model's effectiveness through comparative experiments and ablation studies. Experimental results demonstrate that SentiXRL achieved improved accuracy across multiple standard datasets, showcasing its superior performance in emotion recognition tasks and efficient fine-grained emotion classification in multilingual settings. Compared to traditional discriminative models, SentiXRL is capable of generating richer and more accurate emotion category labels. The Self-circular Analysis Negotiation Mechanism (SANM) further enhances model stability and accuracy through alternating roles of generator and discriminator, facilitating self-supervision and validation. The effectiveness of SentiXRL suggests new directions for future research in emotion recognition.

\section{Limitation}
SentiXRL currently focuses solely on unimodal textual information, and due to limitations in the experimental environment, this study was conducted using a pre-trained model with a maximum of 8 billion parameters. In the future, we will explore the potential for multimodal research and extend our investigations to include testing on Arabic,Hindi and Spanish. These languages, being among the most widely spoken after English and Chinese, will help demonstrate the model's effectiveness across a broader range of linguistic contexts.

\bibliography{coling}

\begin{thebibliography}{43}
\providecommand{\natexlab}[1]{#1}

\bibitem[{Brown et~al.(2020{\natexlab{a}})Brown, Mann, Ryder, Subbiah, Kaplan, Dhariwal, Neelakantan, Shyam, Sastry, Askell, Agarwal, Herbert-Voss, Krueger, Henighan, Child, Ramesh, Ziegler, Wu, Winter, Hesse, Chen, Sigler, Litwin, Gray, Chess, Clark, Berner, McCandlish, Radford, Sutskever, and Amodei}]{NEURIPS2020_1457c0d6}
Tom Brown, Benjamin Mann, Nick Ryder, Melanie Subbiah, Jared~D Kaplan, Prafulla Dhariwal, Arvind Neelakantan, Pranav Shyam, Girish Sastry, Amanda Askell, Sandhini Agarwal, Ariel Herbert-Voss, Gretchen Krueger, Tom Henighan, Rewon Child, Aditya Ramesh, Daniel Ziegler, Jeffrey Wu, Clemens Winter, Chris Hesse, Mark Chen, Eric Sigler, Mateusz Litwin, Scott Gray, Benjamin Chess, Jack Clark, Christopher Berner, Sam McCandlish, Alec Radford, Ilya Sutskever, and Dario Amodei. 2020{\natexlab{a}}.
\newblock \href {https://proceedings.neurips.cc/paper_files/paper/2020/file/1457c0d6bfcb4967418bfb8ac142f64a-Paper.pdf} {Language models are few-shot learners}.
\newblock In \emph{Advances in Neural Information Processing Systems}, volume~33, pages 1877--1901. Curran Associates, Inc.

\bibitem[{Brown et~al.(2020{\natexlab{b}})Brown, Mann, Ryder, Subbiah, Kaplan, Dhariwal, Neelakantan, Shyam, Sastry, Askell, Agarwal, Herbert-Voss, Krueger, Henighan, Child, Ramesh, Ziegler, Wu, Winter, Hesse, Chen, Sigler, Litwin, Gray, Chess, Clark, Berner, McCandlish, Radford, Sutskever, and Amodei}]{brown2020language}
Tom~B. Brown, Benjamin Mann, Nick Ryder, Melanie Subbiah, Jared Kaplan, Prafulla Dhariwal, Arvind Neelakantan, Pranav Shyam, Girish Sastry, Amanda Askell, Sandhini Agarwal, Ariel Herbert-Voss, Gretchen Krueger, Tom Henighan, Rewon Child, Aditya Ramesh, Daniel~M. Ziegler, Jeffrey Wu, Clemens Winter, Christopher Hesse, Mark Chen, Eric Sigler, Mateusz Litwin, Scott Gray, Benjamin Chess, Jack Clark, Christopher Berner, Sam McCandlish, Alec Radford, Ilya Sutskever, and Dario Amodei. 2020{\natexlab{b}}.
\newblock \href {https://arxiv.org/abs/2005.14165} {Language models are few-shot learners}.
\newblock \emph{Preprint}, arXiv:2005.14165.

\bibitem[{Busso et~al.(2008)Busso, Bulut, Lee, Kazemzadeh, Mower, Kim, Chang, Lee, and Narayanan}]{busso2008iemocap}
Carlos Busso, Murtaza Bulut, Chi-Chun Lee, Abe Kazemzadeh, Emily Mower, Samuel Kim, Jeannette~N Chang, Sungbok Lee, and Shrikanth~S Narayanan. 2008.
\newblock Iemocap: Interactive emotional dyadic motion capture database.
\newblock \emph{Language resources and evaluation}, 42:335--359.

\bibitem[{Chen et~al.(2022)Chen, Fan, Xing, Pang, Huang, Han, Tie, and Xu}]{chen2022cped}
Yirong Chen, Weiquan Fan, Xiaofen Xing, Jianxin Pang, Minlie Huang, Wenjing Han, Qianfeng Tie, and Xiangmin Xu. 2022.
\newblock Cped: A large-scale chinese personalized and emotional dialogue dataset for conversational ai.
\newblock \emph{arXiv preprint arXiv:2205.14727}.

\bibitem[{Chung et~al.(2022)Chung, Hou, Longpre, Zoph, Tay, Fedus, Li, Wang, Dehghani, Brahma, Webson, Gu, Dai, Suzgun, Chen, Chowdhery, Castro-Ros, Pellat, Robinson, Valter, Narang, Mishra, Yu, Zhao, Huang, Dai, Yu, Petrov, Chi, Dean, Devlin, Roberts, Zhou, Le, and Wei}]{chung2022scaling}
Hyung~Won Chung, Le~Hou, Shayne Longpre, Barret Zoph, Yi~Tay, William Fedus, Yunxuan Li, Xuezhi Wang, Mostafa Dehghani, Siddhartha Brahma, Albert Webson, Shixiang~Shane Gu, Zhuyun Dai, Mirac Suzgun, Xinyun Chen, Aakanksha Chowdhery, Alex Castro-Ros, Marie Pellat, Kevin Robinson, Dasha Valter, Sharan Narang, Gaurav Mishra, Adams Yu, Vincent Zhao, Yanping Huang, Andrew Dai, Hongkun Yu, Slav Petrov, Ed~H. Chi, Jeff Dean, Jacob Devlin, Adam Roberts, Denny Zhou, Quoc~V. Le, and Jason Wei. 2022.
\newblock \href {https://arxiv.org/abs/2210.11416} {Scaling instruction-finetuned language models}.
\newblock \emph{Preprint}, arXiv:2210.11416.

\bibitem[{Devlin et~al.(2019)Devlin, Chang, Lee, and Toutanova}]{devlin2019bert}
Jacob Devlin, Ming-Wei Chang, Kenton Lee, and Kristina Toutanova. 2019.
\newblock \href {https://arxiv.org/abs/1810.04805} {Bert: Pre-training of deep bidirectional transformers for language understanding}.
\newblock \emph{Preprint}, arXiv:1810.04805.

\bibitem[{Hoang et~al.(2018)Hoang, Nguyen, Le, and Phung}]{2018MGAN}
Quan Hoang, Tu~Dinh Nguyen, Trung Le, and Dinh Phung. 2018.
\newblock Mgan: Training generative adversarial nets with multiple generators.

\bibitem[{Hu et~al.(2023)Hu, Bao, Wei, Zhou, and Hu}]{hu2023supervised}
Dou Hu, Yinan Bao, Lingwei Wei, Wei Zhou, and Songlin Hu. 2023.
\newblock Supervised adversarial contrastive learning for emotion recognition in conversations.
\newblock \emph{arXiv preprint arXiv:2306.01505}.

\bibitem[{Khan and Fu(2021)}]{2021Exploiting}
Zaid Khan and Yun Fu. 2021.
\newblock Exploiting bert for multimodal target sentiment classification through input space translation.

\bibitem[{Lee et~al.(2023)Lee, An, and Thorne}]{lee2023large}
Noah Lee, Na~Min An, and James Thorne. 2023.
\newblock \href {https://arxiv.org/abs/2305.13788} {Can large language models capture dissenting human voices?}
\newblock \emph{Preprint}, arXiv:2305.13788.

\bibitem[{Lei et~al.(2024)Lei, Dong, Wang, Wang, and Wang}]{lei2024instructerc}
Shanglin Lei, Guanting Dong, Xiaoping Wang, Keheng Wang, and Sirui Wang. 2024.
\newblock \href {https://arxiv.org/abs/2309.11911} {Instructerc: Reforming emotion recognition in conversation with a retrieval multi-task llms framework}.
\newblock \emph{Preprint}, arXiv:2309.11911.

\bibitem[{Li et~al.(2021)Li, Lin, Fu, and Wang}]{Li2021PastPA}
JiangNan Li, Zheng Lin, Peng Fu, and Weiping Wang. 2021.
\newblock \href {https://api.semanticscholar.org/CorpusID:244119671} {Past, present, and future: Conversational emotion recognition through structural modeling of psychological knowledge}.
\newblock In \emph{Conference on Empirical Methods in Natural Language Processing}.

\bibitem[{Liu et~al.(2023)Liu, Zhang, Zhang, Xue, and You}]{liu2023hierarchical}
Xiao Liu, Jian Zhang, Heng Zhang, Fuzhao Xue, and Yang You. 2023.
\newblock \href {https://arxiv.org/abs/2305.00262} {Hierarchical dialogue understanding with special tokens and turn-level attention}.
\newblock \emph{Preprint}, arXiv:2305.00262.

\bibitem[{Liu et~al.(2015)Liu, Nourbakhsh, Li, Fang, and Shah}]{liu2015real}
Xiaomo Liu, Armineh Nourbakhsh, Quanzhi Li, Rui Fang, and Sameena Shah. 2015.
\newblock Real-time rumor debunking on twitter.
\newblock In \emph{Proceedings of the 24th ACM International on Conference on Information and Knowledge Management}, pages 1867--1870.

\bibitem[{Liu et~al.(2022)Liu, Yuan, Mao, Liang, Yang, Qiu, Cheng, Li, Xu, and Gao}]{liu2022make}
Yihe Liu, Ziqi Yuan, Huisheng Mao, Zhiyun Liang, Wanqiuyue Yang, Yuanzhe Qiu, Tie Cheng, Xiaoteng Li, Hua Xu, and Kai Gao. 2022.
\newblock \href {https://arxiv.org/abs/2209.02604} {Make acoustic and visual cues matter: Ch-sims v2.0 dataset and av-mixup consistent module}.
\newblock \emph{Preprint}, arXiv:2209.02604.

\bibitem[{Lu et~al.(2019)Lu, Batra, Parikh, and Lee}]{lu2019vilbert}
Jiasen Lu, Dhruv Batra, Devi Parikh, and Stefan Lee. 2019.
\newblock \href {https://arxiv.org/abs/1908.02265} {Vilbert: Pretraining task-agnostic visiolinguistic representations for vision-and-language tasks}.
\newblock \emph{Preprint}, arXiv:1908.02265.

\bibitem[{Mishra et~al.(2022)Mishra, Khashabi, Baral, and Hajishirzi}]{mishra-etal-2022-cross}
Swaroop Mishra, Daniel Khashabi, Chitta Baral, and Hannaneh Hajishirzi. 2022.
\newblock \href {https://doi.org/10.18653/v1/2022.acl-long.244} {Cross-task generalization via natural language crowdsourcing instructions}.
\newblock In \emph{Proceedings of the 60th Annual Meeting of the Association for Computational Linguistics (Volume 1: Long Papers)}, pages 3470--3487, Dublin, Ireland. Association for Computational Linguistics.

\bibitem[{Muhammad et~al.(2022)Muhammad, Adelani, Ruder, Ahmad, Abdulmumin, Bello, Choudhury, Emezue, Abdullahi, Aremu et~al.}]{muhammad2022naijasenti}
Shamsuddeen~Hassan Muhammad, David~Ifeoluwa Adelani, Sebastian Ruder, Ibrahim~Said Ahmad, Idris Abdulmumin, Bello~Shehu Bello, Monojit Choudhury, Chris~Chinenye Emezue, Saheed~Salahudeen Abdullahi, Anuoluwapo Aremu, et~al. 2022.
\newblock Naijasenti: A nigerian twitter sentiment corpus for multilingual sentiment analysis.
\newblock \emph{arXiv preprint arXiv:2201.08277}.

\bibitem[{{\"O}hman et~al.(2020){\"O}hman, P{\`a}mies, Kajava, and Tiedemann}]{ohman2020xed}
Emily {\"O}hman, Marc P{\`a}mies, Kaisla Kajava, and J{\"o}rg Tiedemann. 2020.
\newblock Xed: A multilingual dataset for sentiment analysis and emotion detection.
\newblock \emph{arXiv preprint arXiv:2011.01612}.

\bibitem[{Ouyang et~al.(2022)Ouyang, Wu, Jiang, Almeida, Wainwright, Mishkin, Zhang, Agarwal, Slama, Ray et~al.}]{ouyang2022training}
Long Ouyang, Jeffrey Wu, Xu~Jiang, Diogo Almeida, Carroll Wainwright, Pamela Mishkin, Chong Zhang, Sandhini Agarwal, Katarina Slama, Alex Ray, et~al. 2022.
\newblock Training language models to follow instructions with human feedback.
\newblock \emph{Advances in neural information processing systems}, 35:27730--27744.

\bibitem[{Poria et~al.(2017{\natexlab{a}})Poria, Cambria, Hazarika, Majumder, and Morency}]{2017Context}
Soujanya Poria, Erik Cambria, Devamanyu Hazarika, Navonil Majumder, and Louis~Philippe Morency. 2017{\natexlab{a}}.
\newblock Context-dependent sentiment analysis in user-generated videos.
\newblock In \emph{Proceedings of the 55th Annual Meeting of the Association for Computational Linguistics (Volume 1: Long Papers)}.

\bibitem[{Poria et~al.(2017{\natexlab{b}})Poria, Cambria, Hazarika, Majumder, Zadeh, and Morency}]{poria2017context}
Soujanya Poria, Erik Cambria, Devamanyu Hazarika, Navonil Majumder, Amir Zadeh, and Louis-Philippe Morency. 2017{\natexlab{b}}.
\newblock Context-dependent sentiment analysis in user-generated videos.
\newblock In \emph{Proceedings of the 55th annual meeting of the association for computational linguistics (volume 1: Long papers)}, pages 873--883.

\bibitem[{Poria et~al.(2018)Poria, Hazarika, Majumder, Naik, Cambria, and Mihalcea}]{poria2018meld}
Soujanya Poria, Devamanyu Hazarika, Navonil Majumder, Gautam Naik, Erik Cambria, and Rada Mihalcea. 2018.
\newblock Meld: A multimodal multi-party dataset for emotion recognition in conversations.
\newblock \emph{arXiv preprint arXiv:1810.02508}.

\bibitem[{Rosenthal et~al.(2017)Rosenthal, Farra, and Nakov}]{rosenthal-etal-2017-semeval}
Sara Rosenthal, Noura Farra, and Preslav Nakov. 2017.
\newblock \href {https://doi.org/10.18653/v1/S17-2088} {{S}em{E}val-2017 task 4: Sentiment analysis in {T}witter}.
\newblock In \emph{Proceedings of the 11th International Workshop on Semantic Evaluation ({S}em{E}val-2017)}, pages 502--518, Vancouver, Canada. Association for Computational Linguistics.

\bibitem[{Shen et~al.(2021)Shen, Chen, Quan, and Xie}]{shen2021dialogxl}
Weizhou Shen, Junqing Chen, Xiaojun Quan, and Zhixian Xie. 2021.
\newblock Dialogxl: All-in-one xlnet for multi-party conversation emotion recognition.
\newblock In \emph{Proceedings of the AAAI Conference on Artificial Intelligence}, volume~35, pages 13789--13797.

\bibitem[{Song et~al.(2022)Song, Huang, Xue, and Hu}]{song2022supervised}
Xiaohui Song, Longtao Huang, Hui Xue, and Songlin Hu. 2022.
\newblock Supervised prototypical contrastive learning for emotion recognition in conversation.
\newblock \emph{arXiv preprint arXiv:2210.08713}.

\bibitem[{Tai et~al.(2017)Tai, Yang, Liu, and Xu}]{tai2017memnet}
Ying Tai, Jian Yang, Xiaoming Liu, and Chunyan Xu. 2017.
\newblock Memnet: A persistent memory network for image restoration.
\newblock In \emph{Proceedings of the IEEE international conference on computer vision}, pages 4539--4547.

\bibitem[{Wang et~al.(2016)Wang, Huang, Zhu, and Zhao}]{wang2016attention}
Yequan Wang, Minlie Huang, Xiaoyan Zhu, and Li~Zhao. 2016.
\newblock Attention-based lstm for aspect-level sentiment classification.
\newblock In \emph{Proceedings of the 2016 conference on empirical methods in natural language processing}, pages 606--615.

\bibitem[{Wu et~al.()Wu, Gong, Koo, and Hirschberg}]{wumultimodal}
Zehui Wu, Ziwei Gong, Jaywon Koo, and Julia Hirschberg.
\newblock Multimodal multi-loss fusion network for sentiment analysis.

\bibitem[{Xu et~al.(2019)Xu, Mao, and Chen}]{xu2019multi}
Nan Xu, Wenji Mao, and Guandan Chen. 2019.
\newblock \href {https://doi.org/10.1609/aaai.v33i01.3301371} {Multi-interactive memory network for aspect based multimodal sentiment analysis}.
\newblock In \emph{Proceedings of the AAAI Conference on Artificial Intelligence}, volume~33, pages 371--378.

\bibitem[{Yang et~al.(2022)Yang, Zhao, and Qin}]{yang-etal-2022-face}
Hao Yang, Yanyan Zhao, and Bing Qin. 2022.
\newblock \href {https://aclanthology.org/2022.emnlp-main.219} {Face-sensitive image-to-emotional-text cross-modal translation for multimodal aspect-based sentiment analysis}.
\newblock In \emph{Proceedings of the 2022 Conference on Empirical Methods in Natural Language Processing}, pages 3324--3335, Abu Dhabi, United Arab Emirates. Association for Computational Linguistics.

\bibitem[{Yi et~al.(2022)Yi, Yang, Yuan, Cao, Zhang, and Xiao}]{yi2022contextual}
Jingjie Yi, Deqing Yang, Siyu Yuan, Caiyan Cao, Zhiyao Zhang, and Yanghua Xiao. 2022.
\newblock \href {https://arxiv.org/abs/2207.13254} {Contextual information and commonsense based prompt for emotion recognition in conversation}.
\newblock \emph{Preprint}, arXiv:2207.13254.

\bibitem[{Yingjian et~al.(2023)Yingjian, Jiang, Xiaoping, and Zhigang}]{yingjian2023emotionic}
Liu Yingjian, Li~Jiang, Wang Xiaoping, and Zeng Zhigang. 2023.
\newblock Emotionic: Emotional inertia and contagion-driven dependency modelling for emotion recognition in conversation.
\newblock \emph{arXiv preprint arXiv:2303.11117}.

\bibitem[{Yu et~al.(2023)Yu, Chen, and Xia}]{9765342}
Jianfei Yu, Kai Chen, and Rui Xia. 2023.
\newblock \href {https://doi.org/10.1109/TAFFC.2022.3171091} {Hierarchical interactive multimodal transformer for aspect-based multimodal sentiment analysis}.
\newblock \emph{IEEE Transactions on Affective Computing}, 14(3):1966--1978.

\bibitem[{Yu and Jiang(2019)}]{2019Adapting}
Jianfei Yu and Jing Jiang. 2019.
\newblock Adapting bert for target-oriented multimodal sentiment classification.
\newblock In \emph{Twenty-Eighth International Joint Conference on Artificial Intelligence {IJCAI-19}}.

\bibitem[{Yu et~al.(2020)Yu, Jiang, and Xia}]{2020Entity}
Jianfei Yu, Jing Jiang, and Rui Xia. 2020.
\newblock Entity-sensitive attention and fusion network for entity-level multimodal sentiment classification.
\newblock \emph{IEEE/ACM Transactions on Audio, Speech, and Language Processing}, 28:429--439.

\bibitem[{Yu et~al.(2022)Yu, Wang, Xia, and Li}]{ijcai2022p622}
Jianfei Yu, Jieming Wang, Rui Xia, and Junjie Li. 2022.
\newblock \href {https://doi.org/10.24963/ijcai.2022/622} {Targeted multimodal sentiment classification based on coarse-to-fine grained image-target matching}.
\newblock In \emph{Proceedings of the Thirty-First International Joint Conference on Artificial Intelligence, {IJCAI-22}}, pages 4482--4488. International Joint Conferences on Artificial Intelligence Organization.
\newblock Main Track.

\bibitem[{Zahiri and Choi(2018)}]{zahiri2018emotion}
Sayyed~M Zahiri and Jinho~D Choi. 2018.
\newblock Emotion detection on tv show transcripts with sequence-based convolutional neural networks.
\newblock In \emph{Workshops at the thirty-second aaai conference on artificial intelligence}.

\bibitem[{Zhang et~al.(2023{\natexlab{a}})Zhang, Wang, Yin, Liu, Liu, and Yu}]{zhang2023learning}
Haoyu Zhang, Yu~Wang, Guanghao Yin, Kejun Liu, Yuanyuan Liu, and Tianshu Yu. 2023{\natexlab{a}}.
\newblock Learning language-guided adaptive hyper-modality representation for multimodal sentiment analysis.
\newblock \emph{arXiv preprint arXiv:2310.05804}.

\bibitem[{Zhang et~al.(2023{\natexlab{b}})Zhang, Huang, Ma, Li, Luo, Xie, Qin, Luo, Li, Liu, Guo, and Zhang}]{zhang2023recognize}
Youcai Zhang, Xinyu Huang, Jinyu Ma, Zhaoyang Li, Zhaochuan Luo, Yanchun Xie, Yuzhuo Qin, Tong Luo, Yaqian Li, Shilong Liu, Yandong Guo, and Lei Zhang. 2023{\natexlab{b}}.
\newblock \href {https://arxiv.org/abs/2306.03514} {Recognize anything: A strong image tagging model}.
\newblock \emph{Preprint}, arXiv:2306.03514.

\bibitem[{Zhang et~al.(2023{\natexlab{c}})Zhang, Wang, Li, Dong, Wang, Xian, Li, and Zhang}]{zhang-etal-2023-pay}
Yupeng Zhang, Shensi Wang, Peiguang Li, Guanting Dong, Sirui Wang, Yunsen Xian, Zhoujun Li, and Hongzhi Zhang. 2023{\natexlab{c}}.
\newblock \href {https://doi.org/10.18653/v1/2023.findings-acl.831} {Pay attention to implicit attribute values: A multi-modal generative framework for {AVE} task}.
\newblock In \emph{Findings of the Association for Computational Linguistics: ACL 2023}, pages 13139--13151, Toronto, Canada. Association for Computational Linguistics.

\bibitem[{Zhang et~al.(2021)Zhang, Wang, Li, Liu, Guo, and Yu}]{2021ModalNet}
Zhe Zhang, Zhu Wang, Xiaona Li, Nannan Liu, Bin Guo, and Zhiwen Yu. 2021.
\newblock Modalnet: an aspect-level sentiment classification model by exploring multimodal data with fusion discriminant attentional network.
\newblock \emph{World Wide Web}, (6).

\bibitem[{Zhao et~al.(2023)Zhao, Zhou, Li, Tang, Wang, Hou, Min, Zhang, Zhang, Dong, Du, Yang, Chen, Chen, Jiang, Ren, Li, Tang, Liu, Liu, Nie, and Wen}]{zhao2023survey}
Wayne~Xin Zhao, Kun Zhou, Junyi Li, Tianyi Tang, Xiaolei Wang, Yupeng Hou, Yingqian Min, Beichen Zhang, Junjie Zhang, Zican Dong, Yifan Du, Chen Yang, Yushuo Chen, Zhipeng Chen, Jinhao Jiang, Ruiyang Ren, Yifan Li, Xinyu Tang, Zikang Liu, Peiyu Liu, Jian-Yun Nie, and Ji-Rong Wen. 2023.
\newblock \href {https://arxiv.org/abs/2303.18223} {A survey of large language models}.
\newblock \emph{Preprint}, arXiv:2303.18223.

\end{thebibliography}

\appendix

\appendix

\section{Model Fine-Tuning}

Due to the relatively low proportion of Chinese training data in the original Llama model, we conduct fine-tuning on Chinese instructions to enhance its capability in understanding and expressing Chinese. Before conducting experiments, we fine-tune the model using approximately 3.5 million samples from the BELLE dataset and the moss-003 dataset released by Fudan University's MOSS team. During fine-tuning, we employ the Stanford Alpaca template, an effective training method designed to improve the model's ability to understand and execute instructions more effectively throughout the training process.

\begin{table*}[!hpt]
  \centering
  \caption{Fine-tuning hyperparameter settings for Chinese Instruction}
  \label{tab:training_details}
  \setlength{\tabcolsep}{12pt} 
  \renewcommand{\arraystretch}{1} 
  \begin{tabular}{@{}l>{\centering\arraybackslash}p{5cm}>{\centering\arraybackslash}p{5cm}@{}} \toprule
    & \textbf{Full parameters}  & \textbf{Full parameters} \\ \midrule
    Base Model & Meta-Llama-2-7B-Instruct  & Meta-Llama-3-8B-Instruct \\ \hline
    Epochs & 1 & 2 \\ 
    Learning Rate & 2e-4  & 3e-6 \\ 
    LR Scheduler Type & constant  & cosine \\ 
    Context Length & 2K  & 8K \\ 
    Attention Heads & 32  & 32 \\ 
    Key Value Heads & 32  & 8 \\ 
    Warmup Ratio & 0.01 & 0.1 \\ 
    \bottomrule
  \end{tabular}
\end{table*}

\section{Introduction to Baseline Model}

Here is the detailed introduction of the baseline model in the supplementary experimental section.

\begin{itemize}
  \item MMML
\end{itemize}
\begin{quote}
MMML (Multimodal Multi-loss Fusion Network) proposes a multimodal multi-loss fusion network that compares different fusion methods and evaluates the impact of multi-loss training in multimodal fusion networks. It enhances model performance significantly by integrating contextual information.
\end{quote}

\begin{itemize}
  \item ALMT
\end{itemize}
\begin{quote}
ALMT (Adaptive Language-guided Multimodal Transformer) introduces an Adaptive Hyper-modality Learning (AHL) module to learn representations that suppress irrelevant/conflicting information from visual and audio features under the guidance of language features at different scales. By effectively suppressing redundant information in visual and audio modalities, ALMT improves performance significantly across several popular datasets.
\end{quote}

\begin{itemize}
  \item bcLSTM
\end{itemize}
\begin{quote}
bcLSTM designs a model based on Long Short-Term Memory (LSTM) networks, allowing the capturing of contextual information within the same video segment. By recognizing relationships between environments and characters, it aids in better character emotion classification. This method demonstrates robust generalization capabilities.
\end{quote}

\begin{itemize}
  \item DialogXL
\end{itemize}
\begin{quote}
DialogXL is a pre-trained language model specifically designed for dialogue emotion recognition tasks. By enhancing recursive mechanisms and self-attention mechanisms and improving memory capabilities, DialogXL effectively improves emotion recognition performance. DialogXL modifies XLNet's recursive mechanism from paragraph-level to utterance-level to better simulate dialogue data, and introduces dialogue-aware self-attention to capture useful dependencies within and between speakers.
\end{quote}

\begin{itemize}
  \item CPED(BERT+AVG+MLP)
\end{itemize}
\begin{quote}
The CPED paper introduces the BERT+AVG+MLP model for emotion recognition in dialogues (ERC). This model combines BERT's pre-trained language model with average pooling (AVG) of BERT's hidden layer outputs, passing this output through a multi-layer perceptron (MLP) to predict emotion labels.
\end{quote}

\begin{itemize}
  \item SPCL+CL
\end{itemize}
\begin{quote}
This approach involves a novel Supervised Prototypical Contrastive Learning (SPCL) loss function for emotion recognition in dialogues, outperforming traditional supervised contrastive learning losses. SPCL performs well on imbalanced data categories, is insensitive to training batch size, and reduces computational resource requirements.
\end{quote}

\begin{itemize}
  \item SACL
\end{itemize}
\begin{quote}
SACL (Supervised Adversarial Contrastive Learning) is a supervised adversarial contrastive learning framework. It trains adversarial examples by contrasting perceptual adversarial training to generate worst-case samples. It uses joint category expansion contrastive learning objectives to extract structured representations, effectively leveraging label-level feature consistency and preserving fine-grained intra-class features. To mitigate the negative impact of adversarial perturbations on context-dependent data, the framework includes a context adversarial training strategy to learn more diverse features from contexts, enhancing the model's contextual robustness.
\end{quote}

\begin{itemize}
  \item EmotionIC
\end{itemize}
\begin{quote}
EmotionIC models emotional dependencies in dialogues based on emotional inertia and contagion. Compared to previous ERC models, EmotionIC provides a more comprehensive modeling of dialogues at both feature extraction and classification levels. The model attempts to integrate the advantages of attention-based and recursive-based approaches at the feature extraction level.
\end{quote}

\section{Unified Dataset}

Given the standard baseline datasets chosen, such as CPED, MELD, and EmoryNLP, sourced from dialogue lines of TV show characters, most recorded text sentiments tend towards positive emotions, resulting in an overall imbalanced category distribution. Our unified dataset experiment aims to investigate the impact of this category bias on SentiXRL. Our approach to validation involves gathering a larger-scale, fine-grained emotional data set, aiming not only to balance category proportions but also to mitigate the influence of scripted dialogue styles and situational contexts typical in TV dramas.

\begin{table}[!hpt]
  \centering
  \caption{Twitter Dataset Category Weighting Table}
  \label{tab:training_details}
  \renewcommand{\arraystretch}{1} 
  \begin{tabular}{@{}l>{\centering\arraybackslash}p{1.5cm}>{\centering\arraybackslash}p{1.5cm}>{\centering\arraybackslash}p{1.5cm}@{}} \toprule
    & \textbf{Neutral} & \textbf{Positive}& \textbf{Negative} \\ \midrule
    Twitter 2015 & 0.563 &	1.139	& 2.884 \\ 
    Twitter 2017 & 0.723 & 0.786	& 2.899 \\ 
    \bottomrule
  \end{tabular}
\end{table}

\begin{table*}[h]
  \centering
  \caption{Introduction to Unified Datasets}
  \label{tab:unified-datasets}
  \resizebox{\textwidth}{!}{ 
    \begin{tabular}{@{} l >{\raggedright\arraybackslash}p{5cm} >{\raggedright\arraybackslash}p{5cm} >{\centering\arraybackslash}m{1cm} @{}} 
      \toprule
      \textbf{Dataset} & \textbf{Emotion Categories} & \textbf{Data Scale} \\ 
      \midrule
      OCEMOTION & sadness, happiness, disgust, anger, like, surprise, fear & 35693 \\ 
      Chinese Caption Sentiment Dataset & neutral, happiness, sadness, disgust, anger, surprise, fear & 6583 \\ 
      Smp2020WECT & neutral, happy, angry, sad, fear, surprise & 34768 \\ 
      Smp2020EWECT Covid & neutral, happy, angry, sad, fear, surprise & 13606 \\ 
      Emotion Corpus Microblog & happiness, sadness, disgust, like, fear, surprise, anger & 39661 \\ 
      NLPCC2014 (whole sentence) & happiness, sadness, disgust, like, fear, surprise, anger & 47283 \\ 
      NLPCC2014 & happiness, sadness, disgust, like, fear, surprise, anger & 24715 \\ 
      NLPCC2013 (whole sentence) & happiness, sadness, disgust, like, fear, surprise, anger & 10274 \\ 
      NLPCC2013 & happiness, sadness, disgust, like, fear, surprise, anger & 7915 \\ 
      \bottomrule
    \end{tabular}
  }
\end{table*}

\begin{figure*}[htbp]
  \centering
  \includegraphics[width=1\textwidth]{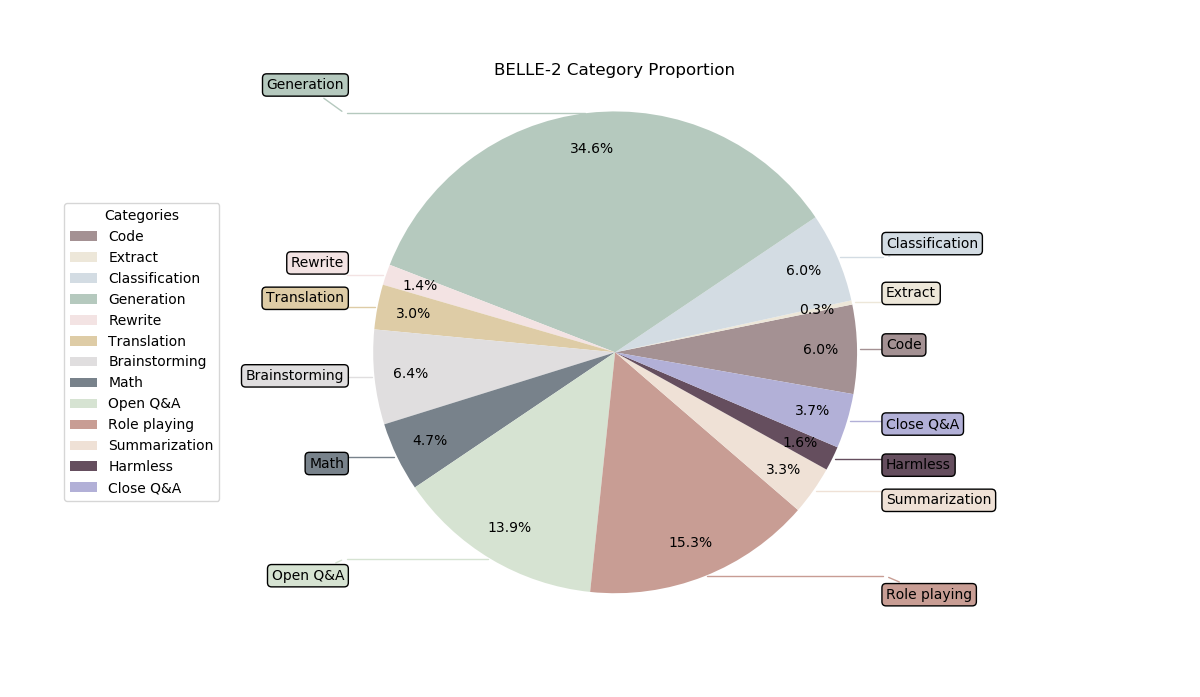}
  \caption{BELLE Data Structure Distribution}
  \label{fig:example}
\end{figure*}

\begin{figure*}[htbp]
  \centering
  \includegraphics[width=1\textwidth]{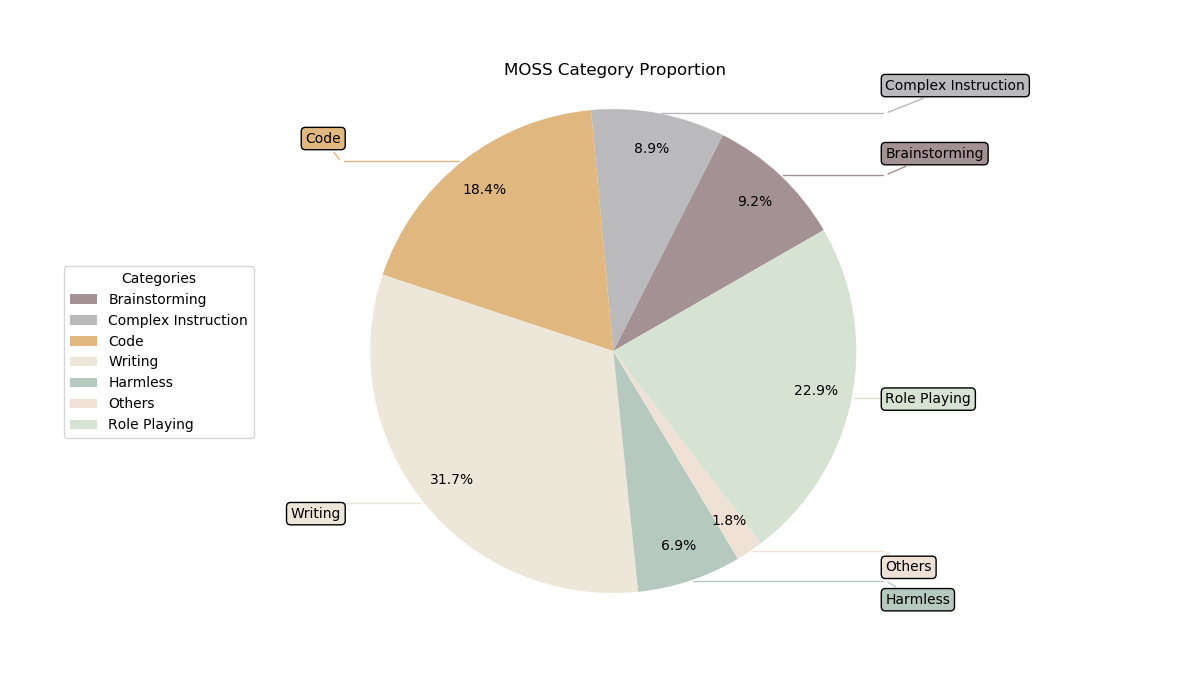}
  \caption{Moss Data Structure Distribution}
  \label{fig:example}
\end{figure*}

\section{Overhead and Computational Efficiency Comparison}
The Llama2 and Llama3 models we used were trained and tested on a single L20 GPU, which indeed requires certain hardware specifications. However, by deploying the model within the SentiXRL framework, we found that on a single L20 GPU, text with a length of less than 500 characters, under the self-circulating analysis and negotiation mechanism (SANM) with $Max\_ round=3$, has an average processing time of 1.8s (based on an average of 1000 sample data). This is not significantly different from the 1.4s for zero-shot deployment of the model alone. Therefore, the additional computational load brought by the SentiXRL framework is within an acceptable range. The computational capacity of large models is indeed highly dependent on the hardware environment, which imposes limitations on deployment.

\end{document}